\documentclass[sigconf]{acmart}
\AtBeginDocument{%
  }

\usepackage{algorithm}
\usepackage{algorithmic}
\usepackage{url}
\usepackage{amsmath}

\usepackage{amssymb}
\usepackage{graphicx}
\usepackage{tabularx}
\usepackage{soul}
\usepackage[dvipsnames]{xcolor}
\definecolor{globalcolor}{RGB}{0, 176, 80}
\definecolor{domaincolor}{RGB}{0, 112, 192}
\usepackage{multirow}
\usepackage{booktabs}
\usepackage{makecell}

\setcopyright{acmlicensed}
\acmConference[WWW '26]{Proceedings of the ACM Web Conference 2026}{April 13-17, 2026}{Dubai, United Arab Emirates}
\copyrightyear{2026}
\acmYear{2026}
\acmDOI{XXXXXXX.XXXXXXX}
\acmISBN{978-1-4503-XXXX-X/2018/06}

\begin{document}

\title[General Anonymizing Multi-Agent System]{GAMA: A General Anonymizing Multi-Agent System for Privacy Preservation Enhanced by Domain Rules and Disproof Mechanism}

\author{Hailong Yang}
\authornote{Equal contribution}
\email{yanghailong@stu.jiangnan.edu.cn}
\orcid{0000-0001-9312-8086}
\affiliation{%
  \institution{Jiangnan University}
  \city{Wuxi}
  \state{Jiangsu}
  \country{China}
}

\author{Renhuo Zhao}
\authornotemark[1]
\email{zhaorenhuo@stu.jiangnan.edu.cn}
\orcid{0009-0000-7475-2676}
\affiliation{%
  \institution{Jiangnan University}
  \city{Wuxi}
  \state{Jiangsu}
  \country{China}
}

\author{Guanjin Wang}
\email{Guanjin.Wang@murdoch.edu.au}
\orcid{0000-0002-5258-0532}
\affiliation{%
  \institution{Murdoch University}
  \city{Murdoch}
  \state{WA}
  \country{Australia}
}

\author{Zhaohong Deng}
\authornote{Corresponding author}
\email{dengzhaohong@jiangnan.edu.cn}
\orcid{0000-0002-0790-6492}
\affiliation{%
  \institution{Jiangnan University}
  \city{Wuxi}
  \state{Jiangsu}
  \country{China}
}

\renewcommand{\shortauthors}{Hailong Yang et al.}

\begin{abstract}
  With the rapid advancement of Large Language Models (LLMs), LLM-based agents exhibit exceptional abilities in understanding and generating natural language, enabling human-like collaboration and information transmission in LLM-based Multi-Agent Systems (MAS). High-performance LLMs are often hosted on web servers in public cloud environments. When tasks involve private data, MAS cannot securely utilize these LLMs without implementing the agentic privacy-preserving mechanism. To address this challenge, we propose a General Anonymizing Multi-Agent System (GAMA), which divides the agents' workspace into private and public spaces, ensuring privacy through a structured anonymization mechanism. In the private space, agents handle sensitive data, while in the public web space, only anonymized data is utilized. GAMA incorporates two key modules to mitigate semantic loss caused by anonymization: Domain-Rule-based Knowledge Enhancement (DRKE) and Disproof-based Logic Enhancement (DLE). We evaluate GAMA on two general question-answering datasets, a public privacy leakage benchmark, and two customized question-answering datasets related to privacy. The results demonstrate that GAMA outperforms existing baselines on the evaluated datasets in terms of both task accuracy and privacy preservation metrics. The project codes of GAMA are available at \url{https://anonymous.4open.science/r/GAMA}.
\end{abstract}



\received{7 October 2025}
\received[revised]{12 March 2009}
\received[accepted]{5 June 2009}

\maketitle

\section{Introduction}
Private data includes sensitive information such as identity, behavior, health, and financial details. The information is highly valuable to individuals, businesses, governments, and society at large \cite{voigt_eu_2017,leiAchievingPersonalizedPrivacypreserving2025}. It enables personalized services, marketing optimization, policymaking, and supports scientific research and public services \cite{martin_role_2017}. However, the misuse of sensitive information may result in identity theft, financial loss, and psychological harm, among other risks \cite{flesca_big_2018, wangUnlearningIncentivizesLearning2025}. Safeguarding privacy is essential to ensure the stable development of society.


Protecting privacy from leakage and misuse is a significant challenge. At the legal level, medical data is regulated by the Health Insurance Portability and Accountability Act (HIPAA) \footnote{HIPAA:\url{https://www.hhs.gov/hipaa/index.html}} in the USA \cite{atchinson_politics_1997}, while the General Data Protection Regulation (GDPR) \footnote{GDPR:\url{https://gdpr-info.eu/}} protects consumer data privacy in the EU \cite{voigt_eu_2017,liaoUnderstandingGDPRNoncompliance2024}.

With the rapid advancement of Large Language Models (LLMs), LLM-based Multi-Agent Systems (MAS) are capable of human-like collaboration and seamless information exchange \cite{sunEffectivenessPrivacypreservingAlgorithms2025}. The knowledge mining and reasoning processes in MAS often rely on powerful LLMs—such as OpenAI's GPT family \cite{achiam_gpt-4_2024}—which are deployed in a public web space accessible to all agents \cite{zhangAgentCFCollaborativeLearning2024}. When processing sensitive data, malicious agents can exploit private information to their advantage \cite{yanUnderstandingDetectingFile2025}. Therefore, privacy preservation has emerged as a pressing and complex agentic challenge in the deployment of MAS. 

To address the privacy challenge in MAS, we propose GAMA—an anonymizing privacy-preserving framework that integrates knowledge- and logic-enhanced mechanisms. GAMA divides the MAS workspace into a private space for processing sensitive data and a public web space for handling anonymized information, ensuring strict privacy boundaries. In the private space, data is anonymized before any interaction with the public web space. To mitigate the impact of anonymization, the public web space features a knowledge-enhanced module based on domain-specific rules and a logic-enhanced module utilizing the disproof mechanism, inspired by the mathematical proof-by-contradiction method. These modules enable reliable reasoning and high-quality decision-making under privacy constraints.

Our contributions are as follows:
\begin{itemize}
    \item  We propose a multi-agent private–public space segregation approach that explicitly separates private and public domains, thereby broadening the scope and applicability of privacy research.
    \item We propose GAMA, a novel and general anonymizing multi-agent system framework that integrates intelligent agents with advanced anonymization techniques and a private–public space separation mechanism to enhance privacy preservation in multi-agent environments.
    \item We introduce two customized datasets tailored for privacy-related question-answering tasks that can be used to facilitate the evaluation of both privacy-preserving performance and task-processing capabilities.
\end{itemize}

The remainder of this paper is structured as follows: Section \ref{sec:releated works} introduces related works, Section \ref{sec:method} details the structure and methodology of GAMA, Section \ref{sec: expriments} presents the experimental results, and Section \ref{sec: conclusion} provides the conclusions and our future works.

\section{Related works}
\label{sec:releated works}

With the rapid advancement of LLMs, LLM-based MAS built upon them have demonstrated improved efficiency in handling complex tasks through holistic modeling and collaborative mechanisms. SPP \cite{wang_unleashing_2024} enhances task performance via multi-turn, multi-persona self-collaboration. AutoAgents \cite{chen_autoagents_2024} proposes an innovative framework that adaptively generates and coordinates multiple specialized agents to form task-oriented AI teams based on specific requirements.

As agents are increasingly applied to replace human labor, privacy concerns have become more prominent. Shi et al. introduced EPEAgents \cite{shi_privacy-enhancing_2025}, which minimizes data flow by sharing only task-relevant and agent-specific information, thereby reducing the risk of privacy leakage. Cui et al. proposed Maris \cite{cui_safeguard-by-development_2025}, a privacy-enhancing development paradigm for MAS, while AgentSafe \cite{mao_agentsafe_2025} protects sensitive information through hierarchical data management.

Privacy is closely linked to self-disclosure, especially involving sensitive attributes like names, social security numbers, and birth dates \cite{adams_anonymate_2019,lukas_analyzing_2023}, which are common in legal \cite{mansfield_behind_2022} and medical texts \cite{yue_phicon_2020}.
Advances in Named Entity Recognition (NER), such as PaDeLLM-NER \cite{lu_padellm-ner_2024} and GPT-NER \cite{wang_gpt-ner_2025}, further improve anonymization accuracy \cite{herwanto_named_2021}. 



\section{Method}
\label{sec:method}

\begin{figure*}[ht]
    \vskip 0.2in
    \begin{center}
        \centerline{\includegraphics[width=1.0\linewidth]{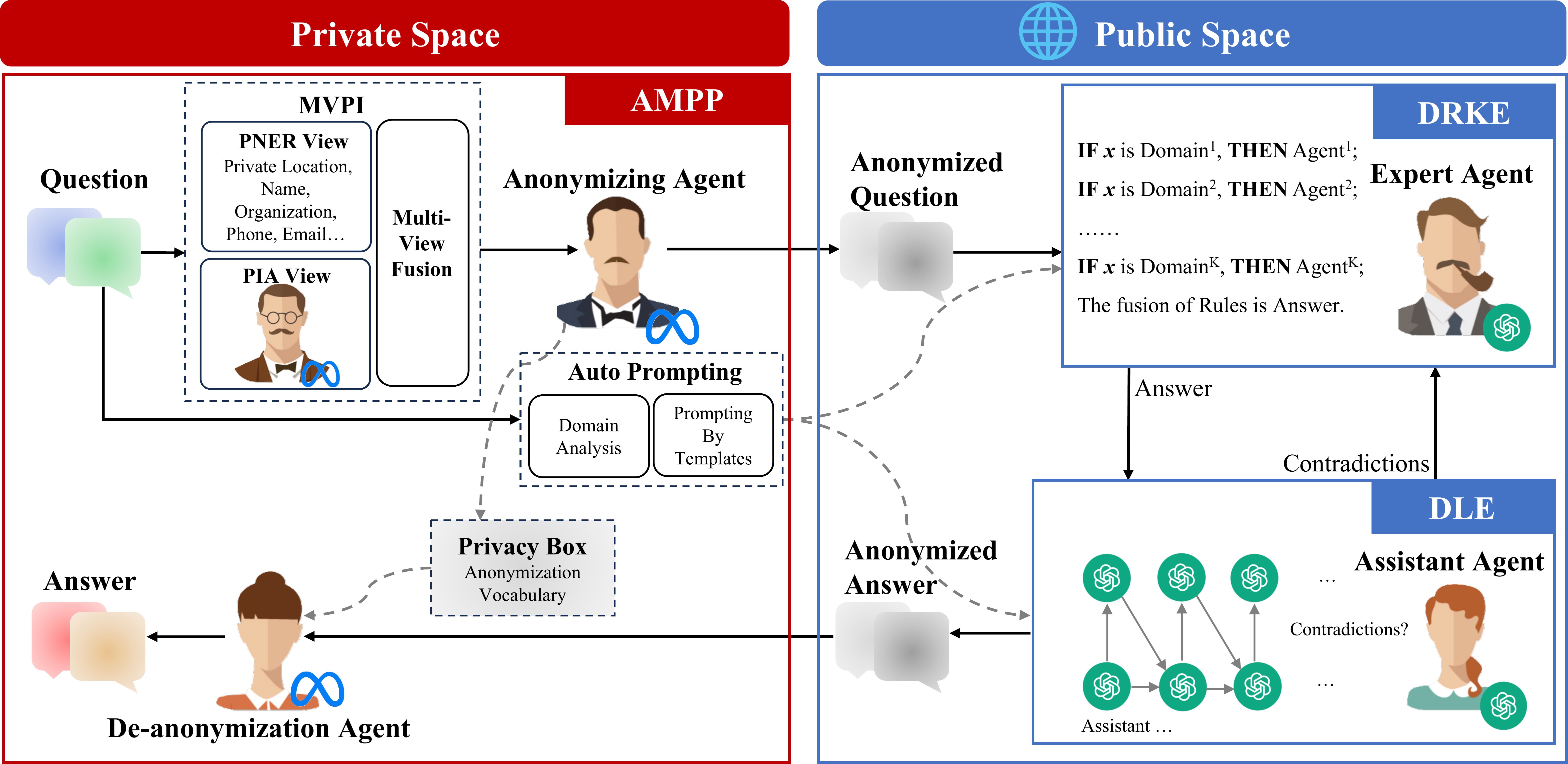}}
        \caption{Structure of GAMA, it has three key parts, including Anonymizing Mechanism for Privacy Preservation (AMPP), Domain-Rule-based Knowledge
        Enhancement (DRKE) and Disproof-based Logic Enhancement (DLE). The Private Space holds privacy-sensitive data and is restricted to entities with high-level permissions, whereas the Public Space remains accessible to all agents under the web environment. The Assistant Agent iteratively generates logical contradictions, while the Expert Agent continuously revises the answer until no further contradictions can be identified.}
        \label{fig:gama}
    \end{center}
    \vskip -0.2in
\end{figure*}

To preserve privacy in MAS, we propose a General Anonymizing Multi-Agent (GAMA) system that enhances privacy preservation through knowledge and logic. GAMA divides the agent workspace into a private space and a public space. The private space is an isolated environment that contains an agent's internal knowledge, data, and reasoning processes, and is inaccessible to others unless explicitly shared. The public space, by contrast, serves as a shared platform for interaction, information exchange, and collaboration, accessible to all authorized agents. GAMA utilizes Anonymizing Mechanism for Privacy Preservation (AMPP) to anonymize sensitive data in private space. GAMA enhances the knowledge and logic of anonymized tasks through Domain-Rule-based Knowledge Enhancement (DRKE), a module designed to recover semantic loss via domain-specific rules, and Disproof-based Logic Enhancement (DLE), a module designed to enhance task logic via the disproof mechanism in the public space, respectively, as illustrated in Figure \ref{fig:gama}. The following subsections detail each key module of GAMA.

\subsection{Anonymizing Mechanism for Privacy Preservation (AMPP)}
AMPP serves as the core module of GAMA, encompassing four essential components: Multi-View Privacy Identification (MVPI), a privacy box, anonymizing and de-anonymization agents, and an auto-prompting module. 

\subsubsection{Multi-View Privacy Identification (MVPI)}
A Multi-View Privacy Identification (MVPI) module is designed based on the method of Named Entity Recognition (NER) \cite{korba_private_2008} and the agent-based method \cite{dou_reducing_2024}  for privacy identification. MVPI incorporates the privacy NER view, the privacy-identifying agent view, and the multi-view privacy fusion module.

\textbf{Privacy NER View }. We have designed the Privacy NER (PNER) view based on the NER-based Privacy Requirements Engineering (NER-PRE) approach \cite{herwanto_named_2021}. It serves to identify privacy-sensitive entities, such as detailed addresses, names, organizations, phone numbers, and emails, etc. The PNER view is a local perspective that identifies private data within the scope of the task, as follows:
\begin{equation}
    \boldsymbol{e}_{v_1}=\mathcal{N}_{priv}(x)\label{eq:pner}
\end{equation}
where $x$ is the natural language token sequence of the task content, $\mathcal{N}_{priv}(.)$ is the PNER model processing function, $\boldsymbol{e}_{v_1}=\{e_1,e_2,…,e_N\}$ is the privacy-named entity set, $e_i$ is the $i$-th privacy-named entity, $i=[1,…, N]$, and $N$ is the total number of named entities.

\textbf{Privacy-Identifying Agent View}. Given the constraints of task-specific content, the PNER view may misclassify non-privacy-named entities as sensitive data. This arises because the privacy attributes of certain named entities depend on contextual semantics and common sense from human society. Effective privacy identification requires a comprehensive assessment from a global perspective. For example, while names of public figures or well-known landmarks might be deemed private from the localized view, such information is publicly accessible and should not be classified as private. To address the issue, we have designed the Privacy-Identifying Agent (PIA) view based on the Reducing Privacy Risk (RPR) approach \cite{dou_reducing_2024}, as follows:
\begin{equation}
    \boldsymbol{e}_{v_2}=\{e_1,e_2,…,e_M\}=\mathcal{A}_{priv}(x) \label{eq:pia}
\end{equation}
where $\mathcal{A}_{priv}(.)$ is the PIA processing function, $\boldsymbol{e}_{v_2}$ is the set of privacy-named entities identified by the agent.

\textbf{Multi-View Fusion}. The PNER view adopts a localized perspective grounded in task-specific content. In contrast, the PIA view offers a global perspective grounded in human social common sense and contextual information. The PNER view identifies all named entities potentially associated with privacy, often resulting in a broader set than the actual ground truth. Conversely, the PIA view leverages rich human knowledge and experience, exhibiting a higher accuracy in identifying common-sense-based privacy. To further improve the identifying accuracy, MVPI fuses the PNER and PIA views through agent-based Multi-View Fusion. The set of privacy-named entities comprises two subsets: the first includes named entities identified by both views. In contrast, the second consists of entities uniquely identified by the fusion module as a complement to the first subset, as illustrated in Eq. (\ref{eq:viewfusion}).
\begin{equation}
    \boldsymbol{e}_{priv} = (\boldsymbol{e}_{v_1}\cap \boldsymbol{e}_{v_2}) \cup  \mathcal{S}_{fus}((\boldsymbol{e}_{v_1}\cup \boldsymbol{e}_{v_2})- (\boldsymbol{e}_{v_2}\cap \boldsymbol{e}_{v_1})) \label{eq:viewfusion}
\end{equation}
where $\cap$ is to compute the intersection of the two sets, $\cup$ is to compute the union of the two sets, $\mathcal{S}_{fus}(.)$ denotes the fusion agent processing function, and $\boldsymbol{e}_{priv}$ is the final output set of privacy-named entities.
\subsubsection{Privacy box: placeholder vocabulary}
After identifying sensitive data, AMPP anonymizes it to prevent potential privacy leakage. The anonymizing process involves replacing privacy-named entities with random placeholders. We have designed the placeholders for AMPP as follow: $<$name$>$, $<$location$>$, $<$organization$>$, $<$phone$>$, $<$email$>$, etc. These placeholders are used to replace private names, addresses, organizations, phone numbers, and email addresses. 

When data migrates from a private space to a public space, the task content must be anonymized. When the data is returned to the private space, the placeholders are converted back to the original sensitive data. To facilitate the bidirectional mapping of privacy preservation, the privacy box has been designed for the mutual mapping between privacy-named entities and placeholders, as shown in Eq. (\ref{eq:vocab}). The sensitive data in the questions is anonymized before leaving the private space. The privacy preservation is achieved by replacing privacy-named entities with placeholders (question anonymization) through the mapping function as shown in Eq. (\ref{eq:map}). Once the answers have been returned to the private space, the de-anonymization agent is responsible for replacing the placeholders with the original privacy-named entities (answer nomination). The anonymized answers are restored, making the full semantic information available to users as shown in Eq. (\ref{eq:remap}). 
\begin{align}
    \mathcal{V}&=Vocab(\boldsymbol{e}_{priv}) \label{eq:vocab}\\
     \boldsymbol{x}^{\prime}&=Map(\boldsymbol{x}| \mathcal{V}) \label{eq:map}\\
     \boldsymbol{y}^{\prime}&=Remap(\boldsymbol{y}|\mathcal{V}) \label{eq:remap}
\end{align}
where $\mathcal{V}$ is the set of mapping relationships between privacy-named entities and placeholders, $\mathcal{V}=\{(e_1,p_1),(e_2,p_2),…,(e_L,p_L)\}$ is the $i$-th mapping relationship, $i=[1,…, L]$, $\boldsymbol{x}^{\prime}$ is the anonymized $\boldsymbol{x}$ and $Map(.)$ is the mapping function from privacy-named entities to placeholders, $\boldsymbol{y}^{\prime}$ is reverse-mapped $\boldsymbol{y}$, and $Remap(.)$ is the function that maps placeholders to privacy-named entities.



\subsubsection{Domain-analysis-based auto prompting}

Auto prompting recognizes target domains of tasks and generates prompts for agents using the presetting templates, as illustrated in Figure \ref{fig:domain disproof} (a). Tasks are classified into various reference domains as shown in \emph{Appendix \ref{app:knowledge domain}}, including \emph{Entertainment, Finance, History}, etc. First, high-order domains are recognized from the semantic level of tasks and human social common sense by the Domain-Analyzing Agent (DAA). By computing the relationship between high-order domains and local reference domains, we constructed an $N*M$ relationship matrix, as illustrated in Eq. (\ref{eq:w2v}), where $N$ represents the number of target domains and $M$ represents the total number of local reference domains. Subsequently, the elementary domain relationship matrix is constructed using the TF-IDF method, as illustrated in Eq. (\ref{eq:tfide}). By adjusting the balance parameter $\alpha$, the high-order relationship matrix is combined with the elementary relationship matrix to generate a new domain membership matrix. Ultimately, the Top-K selecting module is employed to obtain the $K$ domains with the top membership degree, as illustrated in Eq. (\ref{eq:tk}). Auto Prompting automatically generates the relevant agents' prompts according to the target domains and preset templates, and dynamically creates the agents in the public space.
\begin{align}
    \boldsymbol{R}_{ho}&=W2V(DAA(\boldsymbol{x})|\boldsymbol{D}_{ref}) \label{eq:w2v} \\
    \boldsymbol{R}_{elm}&=TFIDF(\boldsymbol{x}|\boldsymbol{D}_{ref}) \label{eq:tfide} \\
    \boldsymbol{D}_{tar}&=\mathcal{T}_K (\alpha \boldsymbol{R}_{ho}+(1-\alpha)\boldsymbol{R}_{elm}) \label{eq:tk}
\end{align}
where $\boldsymbol{D}_{ref}$ is the set of local reference domains, $DAA(.)$ is the DAA processing function that outputs $N$ domains, $\boldsymbol{R}_{ho}$ is the high-order relationship matrix between the high-order domains and the local reference domains, $\boldsymbol{R}_{elm}$ is the elementary relationships between the task content and the local reference domains, $\boldsymbol{R}_{ho}\in \mathbb{R}^{(N*M)}$, $\boldsymbol{R}_{elm}\in \mathbb{R}^{1*M}$, and $\boldsymbol{D}_{tar}$ is target domains of tasks, $\boldsymbol{D}_{tar}\in \mathbb{R}^{1*K}$.

\begin{figure*}[ht]
\vskip 0.2in
\begin{center}
\centerline{\includegraphics[width=1.0\linewidth]{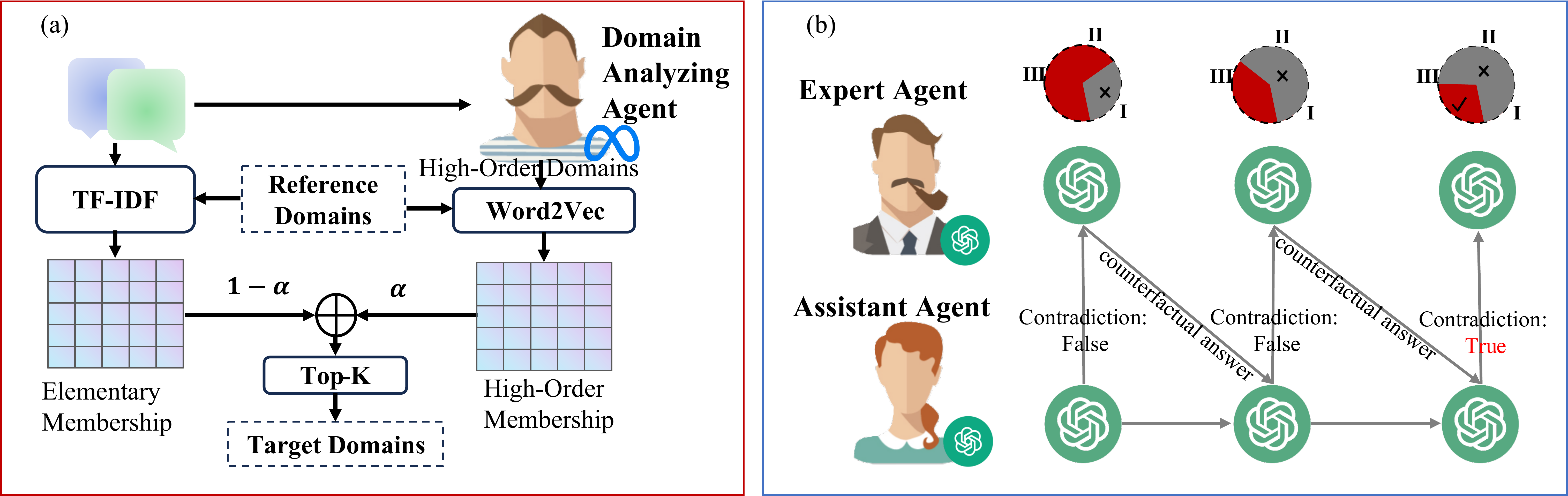}}
\caption{Domain analysis and disproof-based inference, (a) is to analyze the domains of the task for DRKE in Section \ref{sec:drke}, (b) is the disproof process for DLE in Section \ref{sec:dle}.}
\label{fig:domain disproof}
\end{center}
\vskip -0.2in
\end{figure*}

\subsection{Domain-Rule-based Knowledge Enhancement (DRKE)}
\label{sec:drke}
Anonymizing task content can result in the loss of semantic information. To mitigate this, GAMA introduces the Domain-Rule-based Knowledge Enhancement (DRKE) module. DRKE first identifies the relevant domains using auto-prompting, then automatically constructs domain-specific rules. These rules are applied to infer answers across domains, which are finally aggregated to produce the final output.

The construction of domain rules comprises two principal steps: the generation of the antecedents (IF-Parts) and the design of the consequents (THEN-Parts). IF-Parts are term-based membership calculations. The terms related to the domain membership are \{\emph{Extremely Low,  Low, Medium, Moderately High, High, Extremely High}\}. The membership calculation is carried out through semantically reasoning based on local LLMs. IF-Parts focus on semantically related domains. THEN-Parts are the Domain Expert Agents (DEAs) for task processing. DEAs are constructed based on the prompts by Auto Prompting and the preset prompt templates. The outputs of DEAs are the professional answers to various domains.

The memberships of the $K$ domain rule antecedents, as illustrated in Eq. (\ref{eq:rule}), and the outputs of the consequents are jointly involved in the generation of the final answer. The entire fusion process is also LLM-based, with semantically reasoning. The fusing agent conducts semantically reasoning and fuses rule outcomes. The adoption of multi-rule strategies enables a more comprehensive and effective utilization of the knowledge potential embedded within LLM. Furthermore, it facilitates the effective suppression of potential hallucinations in LLM through multi-rule semantic fusion.
\begin{equation}
    \begin{cases}
        \textbf{IF}\, \boldsymbol{x}\, is\, Ent,\, \textbf{THEN}\, \boldsymbol{y}^{1}=DEA^{1}(\boldsymbol{x});\\
        \textbf{IF}\, \boldsymbol{x}\, is\, Fin,\, \textbf{THEN}\, \boldsymbol{y}^{2}=DEA^{2} (\boldsymbol{x});\\
        ...\\
        \textbf{IF}\, \boldsymbol{x}\, is\, His,\, \textbf{THEN}\, \boldsymbol{y}^K=DEA^{K}(\boldsymbol{x}); \label{eq:rule}
    \end{cases}
\end{equation}
where $is$ denotes the membership degree operator, $Ent$ represents the entertainment domain, $Fin$ represents the financial domain, and $His$ represents the historical domain.

\subsection{Disproof-based Logic Enhancement (DLE)}
\label{sec:dle}



One method of disproof in mathematics is proof by contradiction. The process commences with the assumption that a proposition is true. Subsequently, an apparent contradiction is identified, which leads to the conclusion that the original assumption is false and the original proposition is thereby proven \cite{epp_unified_1998}. For the QA task, the proof by contradiction is presented as follows:

\begin{equation}
    \begin{aligned}
        \Gamma,Q,\lnot A \rightarrow \bot \equiv Q \rightarrow A \label{eq: disproof}
    \end{aligned}
\end{equation}
where $\Gamma$ is the context, $Q$ is the question, $\lnot A$ is the counterfactual (or assumed-to-be-false) answer, and $\bot$ denotes a contradiction.

The anonymization process can compromise the integrity of the task logic. To mitigate this, GAMA incorporates the Disproof-based Logic Enhancement (DLE) module to reinforce logical consistency. DLE applies iterative disproof reasoning until the assistant agent detects a contradiction, then outputs the final answer. As shown in Figure~\ref{fig:domain disproof}(b), the workflow operates as follows: the expert agent first proposes a counterfactual answer to the question, and the assistant agent evaluates whether a contradiction exists between them. Based on the assistant's judgment, the expert agent performs further reasoning and generates a revised counterfactual answer. This process is repeated until the assistant agent detects a contradiction, with the final output being the last confirmed answer.


DLE prompts the LLM to engage in continuous logical reasoning and self-correction by identifying contradictions in the propositions. DLE enhances the logical consistency of tasks and answers, effectively suppressing LLM's hallucinations.

\section{Experiments}
\label{sec: expriments}
\subsection{Preliminaries}


In the public web space, agents utilize OpenAI’s GPT-4. In the private environment, local agents use Llama3-8B on a local machine, with hyperparameters detailed in \emph{Appendix \ref{app:llama}}. For text similarity, we use a BERT-based model with 12 Transformer layers, 768 hidden units, and 12 attention heads. The privacy identification model, PNER, is fine-tuned from the pre-trained bert-large-NER \cite{tjong_kim_sang_introduction_2003} on our private datasets. The experiments were performed on a server equipped with an Intel 8352V CPU (54M Cache, 2.10 GHz, 36 cores), 256 GB of RAM, and six NVIDIA A6000 GPUs with 48 GB of GPU RAM (VRAM) each.

To evaluate privacy preservation, we use Precision, Recall, and F1-Score for privacy identification. BLEU \cite{papineni_bleu_2002} measures the fluency of questions after privacy protection, while BERT-based Similarity \cite{tracz_bert-based_2020} assesses their semantic consistency with the original. Leakage Rate (LR) \cite{shao_privacylens_2024} quantifies how much sensitive information is leaked through model outputs. Following \cite{wang_unleashing_2024}, we also use an automatic Score metric to detect factual errors and assess the model’s ability to integrate knowledge across domains. The corresponding equations are provided in \emph{Appendix \ref{app:metrics}}.


We compare GAMA with Standard-Prompting (Standard, one-turn QA agent) \cite{wang_unleashing_2024}, general MAS baselines (SPP, AutoAgents), and privacy-enhancing baselines (EPEAgents, MarisAgents, AgentSafe), where MarisAgents are MAS developed based on the Maris paradigm. In terms of privacy identification capability comparison, we used NER-PRE \cite{herwanto_named_2021}, RPR \cite{dou_reducing_2024}, AgentSafe, and our GAMA.

\begin{figure}
    \centering
    \includegraphics[width=1.0
    \linewidth]{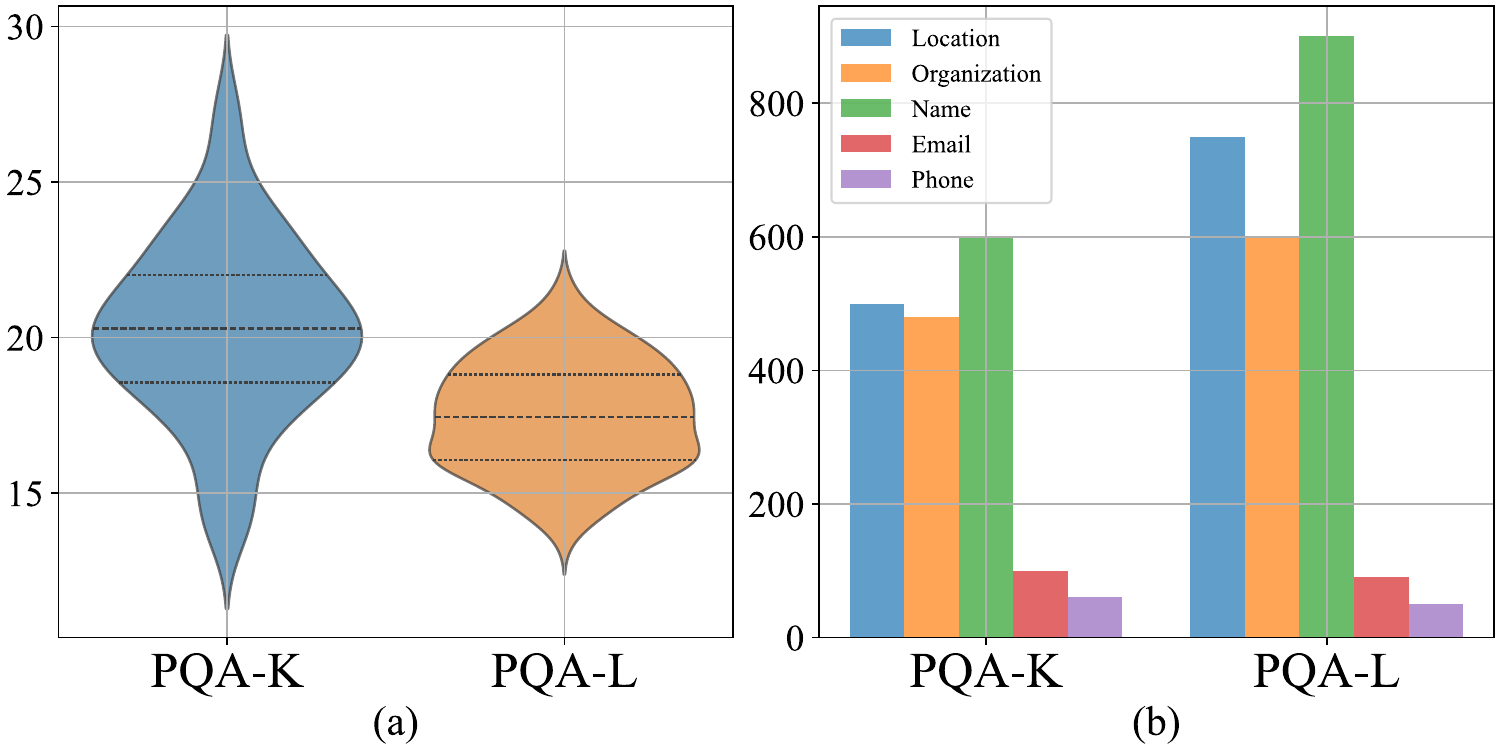}
    \caption{Privacy named entity number distribution of PQA-K and PQA-L. (a) is the privacy named entity distribution of each sample. (b) is the statistics of privacy-named entities from different categories.}
    \label{fig:dataset distribution}
\end{figure}

We adopt two general Question-Answering (QA) datasets, Trivia Creative Writing (TCW) and Logic Grid Puzzle (LGP) \cite{wang_unleashing_2024} to evaluate the general QA performance. To evaluate the capability of preventing privacy leakage, we introduce Privacy Lens (PLens) benchmark \cite{shao_privacylens_2024}. TCW, LGP, and PLens are detailed in \emph{Appendix \ref{app: dataset details}}. To jointly assess both QA performance and privacy preservation ability, we construct two novel GDPR-guided QA datasets—PQA-K and PQA-L. Fig. \ref{fig:dataset distribution} presents the distribution of privacy-related named entities. The two customized datasets are detailed below:
\begin{itemize}
     \item \textbf{PQA-K}. We designed the Privacy QA knowledge (PQA-K) dataset for the knowledge-typed task with private data. We use PQA-K to assess task performance under privacy-preserving conditions. First, the models for evaluation must identify and anonymize private data embedded in questions. Subsequently, the models answer 5 knowledge-typed questions inspired by HotpotQA \cite{yang_hotpotqa_2018}, crafting coherent responses as stories or letters. PQA-K includes 100 instances,  each with five questions, totaling 500 knowledge-typed questions. 
    \item \textbf{PQA-L}. The Privacy QA Logic (PQA-L) is designed to evaluate the ability to process de-identified text and perform logical reasoning. The models for evaluation identify and anonymize private data embedded in paragraphs and questions. The models answer the anonymized questions. The Boolean logical reasoning questions of PQA-L are derived from BoolQ \cite{clark_boolq_2019}. Each instance involves True/False questions based on paragraph content. And each question requires logical inference and text comprehension. PQA-L includes 150 instances with Boolean questions.
\end{itemize}

\begin{figure}
    \centering
    \includegraphics[width=1\linewidth]{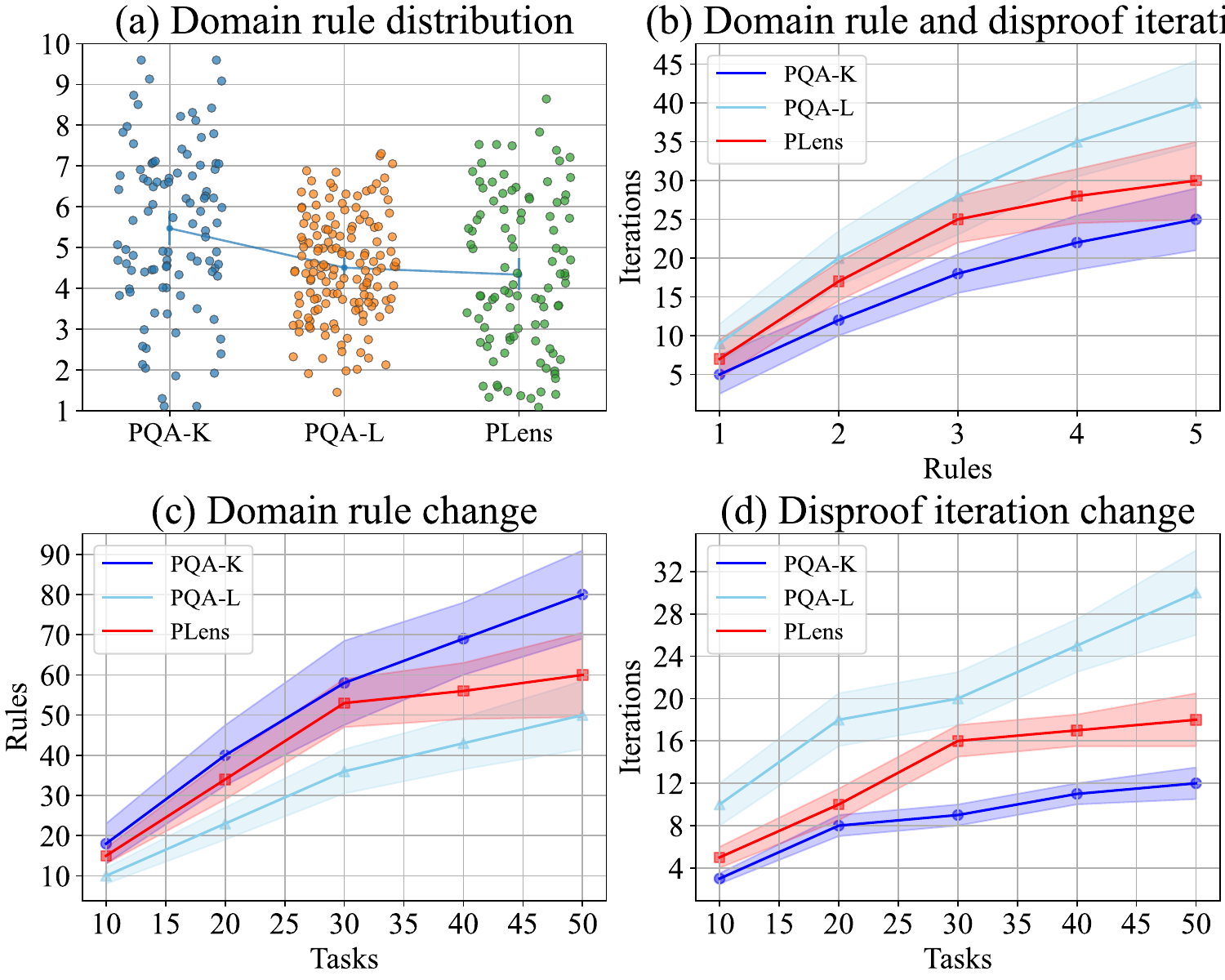}
    \caption{Domain rules and disproof iteration.}
    \label{fig:rule and iteration}
\end{figure}

\begin{figure}[ht]
\begin{center}
\centerline{\includegraphics[width=1\linewidth]{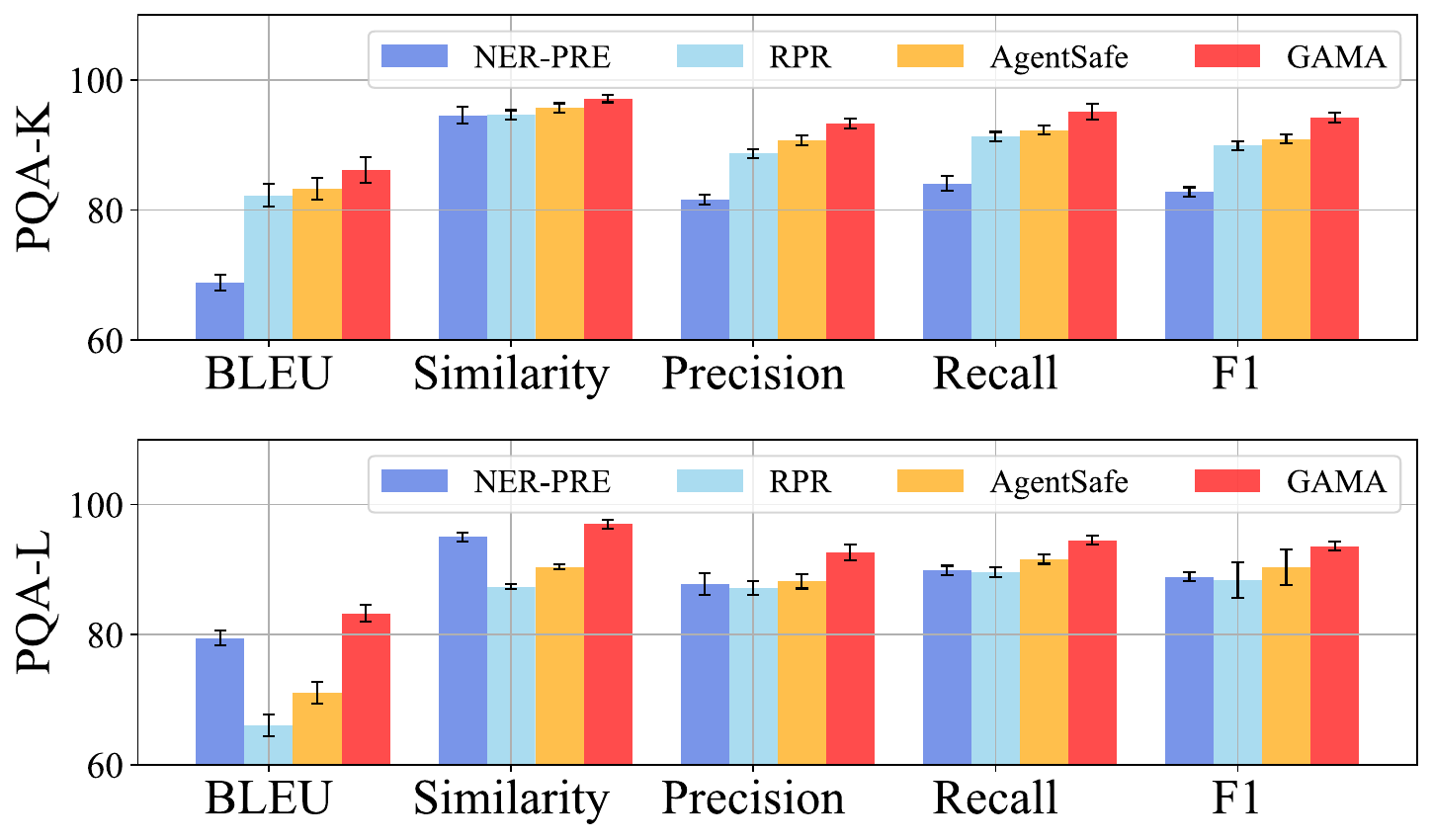}}
\caption{The performance comparison of privacy identification.}
\label{fig:priv identify}
\end{center}
\end{figure}

\subsection{Privacy preservation analysis}


\begin{table}[t]
    \caption{Performance comparison using GPT-4o.}
    \label{tab:privqa}
    \setlength{\tabcolsep}{1.2mm}
    \centering
    \begin{tabular}{lcccccc}
        \toprule[2pt]
        \multirow{2}{*}{Methods} &
        Score$\uparrow$& 
        Score$\uparrow$& 
        LR$\downarrow$ &
        Score$\uparrow$& 
        Score$\uparrow$\\
        &@TCW &
        @LGP &
        @PLens &
        @PQA-K & 
        @PQA-L\\
        \midrule[1pt]
        Standard &     73.1& 54.7& 26.4&39.1& 	44.4\\
        SPP &	       78.5& 67.9&  26.1&45.7&  66.1\\
        AutoAgents&    81.9& 70.7&  26.2&50.8&	71.4\\
        EPEAgents&     -   & -   &  24.3&52.7&	76.1\\
        MarisAgents&   -   & -   &  24.8&53.5&	79.5\\
        AgentSafe&	   -   & -   &  22.5&54.1&	81.1\\
        GAMA(ours) &  \textbf{83.3}& \textbf{74.8}& \textbf{21.7}&\textbf{54.8} &\textbf{82.0} \\
        \bottomrule[2pt]
    \end{tabular}
\end{table}

To assess both general QA performance and privacy-preserving capability, we conducted experiments on TCW, LGP, the authoritative privacy leakage benchmark (PLens), and our two customized datasets (PQA-K and PQA-L). As shown in Table \ref{tab:privqa}, GAMA consistently achieves the highest question-answering performance across both knowledge-based and logic-based questions involving private information, while also exhibiting the lowest privacy leakage.

The results further reveal the following observations:
\begin{itemize}
    \item In general QA tasks, GAMA outperforms AutoAgents, achieving a score improvement of 1.4 points on TCW and 3.9 points on LGP. Since the general QA datasets contain no private data, the latest privacy-preserving MAS methods are not applicable and therefore do not have scores.
    \item Across the privacy QA tasks, the performance gains of MAS methods over the Standard baseline are more substantial on logic-intensive questions.
    \item Notably, SPP and AutoAgents, which lack privacy-preserving mechanisms, have leakage rates similar to Standard and much higher than those of EPEAgents, MarisAgents, AgentSafe, and GAMA, confirming the effectiveness of privacy preservation.
\end{itemize}

To further validate GAMA’s handling mechanisms across tasks, we examined its use of domain rules and disproof iterations. Fig. \ref{fig:rule and iteration}(a) shows that the distribution of domain rules differs by task type. Fig. \ref{fig:rule and iteration}(b) reveals that knowledge tasks employ more rules yet fewer disproof iterations, whereas logic tasks display the opposite pattern. This indicates GAMA augments knowledge processing via domain rules and strengthens logical reasoning through disproof. Fig. \ref{fig:rule and iteration}(c) and (d) track how rule counts and iterations evolve with tasks: logic tasks generally incur more disproof iterations. In contrast, knowledge tasks involve more domain rules. Thus, GAMA flexibly adapts its strategy to each task type, thereby improving performance.

In summary, GAMA outperforms all baseline methods in terms of general QA, privacy leakage, and privacy QA.

\subsection{Privacy-identifying analysis}
To verify the privacy identification capabilities of GAMA, we evaluated four methods: NER-PRE, RPR, AgentSafe, and our GAMA. As shown in Figure \ref{fig:priv identify}, GAMA consistently outperforms all baselines in identifying privacy across both knowledge and logic domains. RPR outperforms NER-PRE on knowledge-type tasks due to its LLM-based design, which captures broader contextual and societal knowledge; however, it falls behind on logic-type tasks. AgentSafe ranks second overall by categorizing information flow based on safety rankings, outperforming both NER-PRE and RPR on most metrics. AgentSafe relies on LLMs and expert knowledge, lacking NER support. By integrating both PNER and PIA views, GAMA achieves superior performance in identifying both knowledge-based and logic-based privacy.

\subsection{Re-identification attack test}
GAMA is a comprehensive privacy protection framework designed to safeguard against adversarial attacks, particularly re-identification attacks \cite{cho_generalizable_2024}, while maintaining robust privacy defenses. To demonstrate GAMA's resilience against adversarial threats, we conducted a re-identification attack test using ARX\footnote{ARX is a data anonymization tool, \url{Data Anonymization Tool}}, a data anonymization tool, with the results presented in Table \ref{tab:arx}. The results from experiments across three attack scenarios—prosecutor, journalist, and marketer—show that re-identification attacks on GAMA-anonymized named entities consistently have a success rate lower than $0.21\%$. The experimental results highlight the framework's strong resilience to adversarial threats \cite{el_emam_guide_2020}.

\begin{table}[t]
    \caption{Re-identification Attack Test}
    \label{tab:arx}
    \setlength{\tabcolsep}{1.2mm}
    \centering
    \begin{tabular}{lcccccc}
        \toprule[2pt]
        Attacker &Record at Risk& Highest Risk & Success rate\\
        \midrule
        prosecutor&	   0.30\%&	0.30\%& 0.00\%\\
        journalist&	   0.83\%&	1.12\%& 0.21\%\\
        marketer&      0.40\%&	1.00\%& 0.01\%\\
        \bottomrule[2pt]
    \end{tabular}
\end{table}

\begin{figure*}
    \centering
    \includegraphics[width=1.0\linewidth]{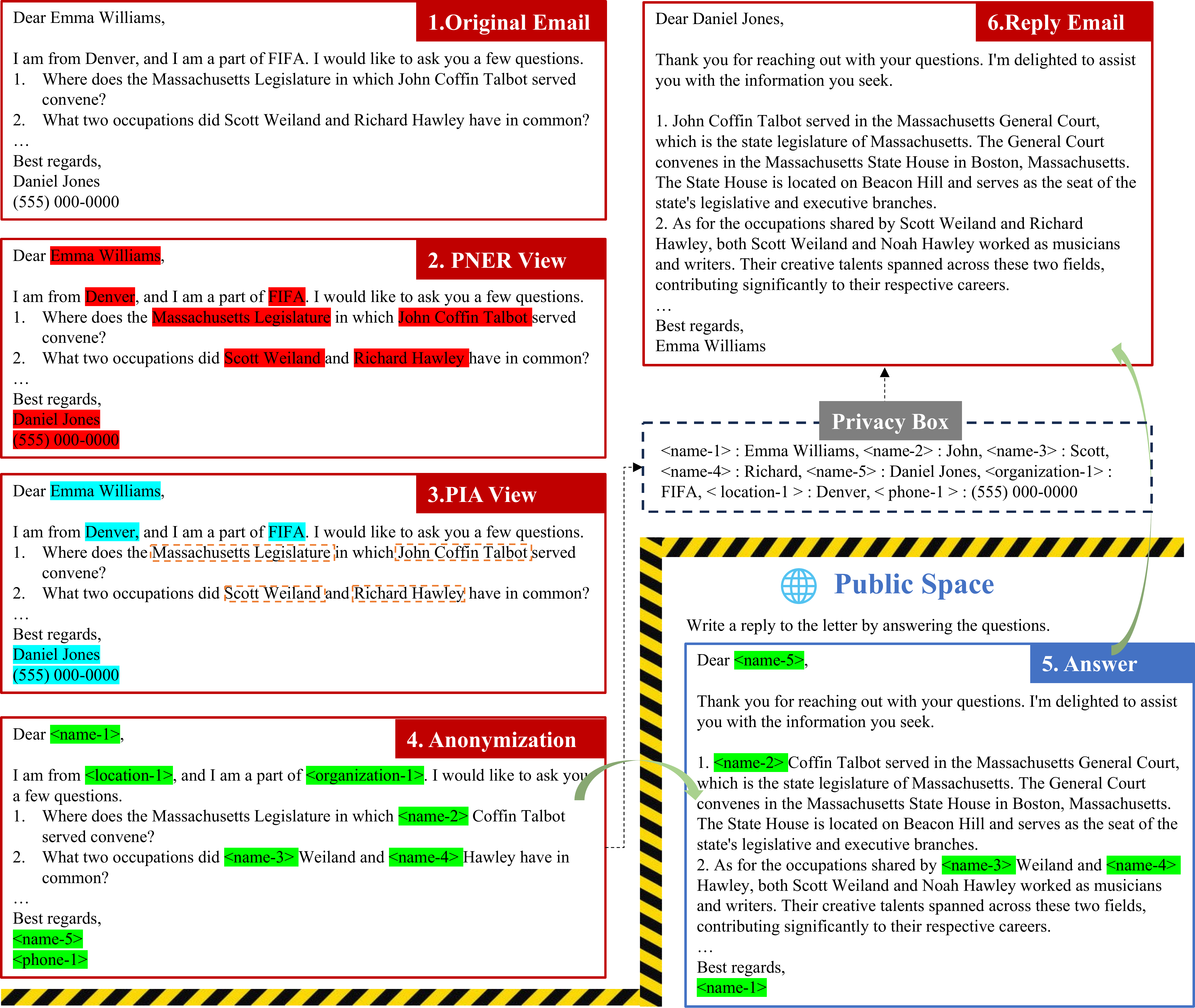}
    \caption{Case study: automatic email reply using GAMA.}
    \label{fig:email}
\end{figure*}

\subsection{Case study: automatic email reply}  

To further evaluate GAMA, we applied it to email processing. As shown in Figure \ref{fig:email}, GAMA replied to an email that contained private information like names, phone numbers, and organizations.


GAMA first detects private information using the PNER and PIA views. The main difference lies in four named entities: \emph{Massachusetts Legislature, John Coffin Talbot, Scott Weiland, and Richard Hawley}. PNER classifies them as private based on entity type, while PIA, guided by common sense, treats them as public figures and non-private. After combining both views, GAMA continues to randomly anonymize these names to protect contextual privacy. The anonymizing agent builds a privacy box to map placeholders back to the original sensitive entities.


In the public space, GAMA enhances logic and knowledge. For knowledge, DRKE uses domain rules from politics and art to extract and complete information about figures mentioned in emails. For logic, DLE applies the disproof mechanism to reason step by step—for example, locating the \emph{Massachusetts legislature} through \emph{General Court, Boston}, and finally \emph{Beacon Hill}. GAMA then generates a reply email with placeholders. In the private space, the de-anonymization agent remaps the placeholders to their original values using the privacy box. As shown in the figure, the final email contains no anonymized placeholders.

The case study is further evidence that GAMA is effective in identifying privacy, anonymizing it, and finally safeguarding the privacy in the original text.

\subsection{Compute source analysis}

GAMA runs several models locally within a private environment, including Llama3 8B, BERT-Large for NER, and a Word2Vec embedding model, collectively consuming approximately 20GB of GPU memory. GAMA primarily utilizes substantial GPU memory, while its RAM consumption remains relatively low. Each question-answering session takes more than two minutes on average.

We further evaluated GAMA's performance on the PLens dataset and compared it with other baselines, as presented in Table \ref{tab:compute complexity}. The average response time per query for GAMA is 118 seconds, making it the most efficient method among MAS methods in terms of time complexity. Regarding space complexity, GAMA requires 370.16 MB less memory and 6 points less CPU utilization than MarisAgents.Since GAMA runs multiple models locally, it is the only method that consumes GPU resources, while all other baselines exhibit zero GPU memory usage. Moreover, because part of GAMA’s data processing is performed in the local private space via GPU computation, it requires the least number of public web LLM API tokens—0.1K fewer than the lowest baseline, SPP.

In summary, GAMA attains the lowest time and space complexity among all privacy-preserving multi-agent systems.


\begin{table}[t]
    \caption{Computational Complexity on PLens using GPT-4o.}
    \label{tab:compute complexity}
    \setlength{\tabcolsep}{1.5mm}
    \centering
    \begin{tabular}{lccccc}
        \toprule[2pt]
        Methods &Time &RAM &CPU & GPU RAM &Tokens\\
        \midrule
        SPP&	          125s&	   27.22 MB&	  9\%&    -&     1.9K\\
        AutoAgents&	      156s&	    28.22 MB&	  11\%&    -&     2.5K\\
        EPEAgents&         200s&    400.11 MB&   13\%&      -&     3.1K\\
        MarisAgents&       259s&    500.41 MB&    14\%&      -&    6.3K\\
        AgentSafe&         198s&    420.90 M&     13\%&      -&     4.5K\\
        GAMA(ours)&	       118s&	130.25 MB&    8\%&    20GB&     1.8K\\
        \bottomrule[2pt]
    \end{tabular}
\end{table}

\subsection{Ablation study}
\begin{table}[t]
    \caption{Ablation Study of GAMA.}
    \label{tab:ablation}
    \setlength{\tabcolsep}{1.5mm}
    \centering
    \begin{tabular}{lccccc}
        \toprule[2pt]
        \multirow{2}{*}{Methods} &
        Score$\uparrow$& 
        Score$\uparrow$&
        LR$\downarrow$&
        Score$\uparrow$& 
        Score$\uparrow$\\
        &@TCW &
        @LGP &
        @PLens&
        @PQA-K & 
        @PQA-L\\
        \midrule
        GAMA	&83.3& 74.8&21.7 &54.8	&82.0 \\
        \,-AMPP	&\textbf{83.5} &\textbf{74.9} &\textbf{20.1}& \textbf{61.8}	&\textbf{84.0} \\
        \,-DRKE	&82.7& 71.1 &22.0 &57.3	&82.5 \\
        \,-DLE	&80.3& 69.5 &24.7 &53.2	&76.9 \\
        \bottomrule[2pt]
    \end{tabular}
\end{table}

To analyze the effectiveness of multiple modules in GAMA, we conduct an ablation study. In Table \ref{tab:ablation}, the suffix “-AMPP” indicates the disabling of the AMPP mechanism within GAMA, “-DRKE” signifies the exclusion of domain rules for enhancing the knowledge mining ability of the expert agent in the public space, and “-DLE” denotes the absence of the disproof-based logic enhancement module. 

From the data in Table \ref{tab:ablation}, we find that GAMA’s performance increases by 7 points on PQA-K and 2 points on PQA-L when AMPP is turned off. This improvement can be attributed to the potential loss of semantic information during AMPP’s anonymization process. It's worth noting that the loss of the logic-typed PQA-L is less than the loss of the knowledge-typed PQA-K. Through the analysis of samples, we find that the identified parts of privacy are mostly knowledge-typed, and placeholders do not replace the logic parts of questions after anonymization. After disabling the DRKE module, the performance of GAMA is reduced, and this degradation is particularly pronounced on the PQA-K dataset. When the DLE module is disabled, there are no enhancement modules, and GAMA's performance is lowest both on PQA-K and PQA-L. In conclusion, the AMPP, DRKE, and DLE modules are crucial to the overall performance of GAMA. 

\subsection{Significance test}
We conducted a post-hoc significance analysis of GAMA using the Friedman test \cite{friedman_bias_1997}. The average ranks obtained by each method are reported as part of the test results, along with the p-values obtained by applying post hoc methods to the results of the Friedman procedure. The Friedman statistic (chi-square distribution with 5 degrees of freedom) is 22.485714, and the corresponding p-value is 0.000423. The p-value is much smaller than the significance level $\alpha=0.05$, indicating that GAMA exhibits statistically significant differences compared to the baseline methods.

Holm's \cite{holm_simple_1979} procedure rejects those hypotheses that have an unadjusted p-value $\le0.025$. Table \ref{tab:holm test} shows that our GAMA rejects all the hypotheses. Overall, GAMA has statistically significant differences compared to all the baselines. 


\begin{table}[!htp]
\caption{Post Hoc comparison}
\label{tab:holm test}
\setlength{\tabcolsep}{2mm}
\centering
    \begin{tabular}{clcccc}
        \toprule[2pt]
        $i$ &Algorithm      &$z$ &p-value    &Holm       &Hypothesis\\
        \midrule
        6   &   EPEAgents   &3.7187         &0.0001    &0.0008       &Reject\\
        5   &   Standard    &3.2116         &0.0005     &0.0013     &Reject\\
        4   &   MarisAgents &2.2007         &0.0017    &0.0017   &Reject\\
        3   &   SPP         &1.6903         &0.0082   &0.0025      &Reject\\
        2   &   AgentSafe   &1.1904         &0.0124   &0.0075     &Reject\\
        1   &   AutoAgents  &0.8451         &0.0217   &0.0125       &Reject\\
        \bottomrule[2pt]
    \end{tabular}
\end{table}


\section{Conclusion}
\label{sec: conclusion}
LLM-based Multi-Agent System (MAS) encounters privacy risks. To address the issue, we propose a General Anonymizing Multi-Agent system, GAMA. The AMPP mechanism must anonymize the tasks in GAMA before being transmitted to the public space. The agents in the public space are only permitted to tackle anonymized tasks. To address the issue of semantic loss resulting from anonymization, GAMA incorporates DRKE and DLE enhancement modules. 

Experiments were conducted on two general QA datasets and three privacy QA datasets, including our custom-built PQA-K and PQA-L. The results show that GAMA surpasses state-of-the-art models in task performance while also exhibiting clear advantages in privacy preservation. Overall, GAMA demonstrates consistent superiority in both task effectiveness and privacy protection across multiple domains.

However, several limitations need to be addressed. First, the criteria employed by GAMA for privacy identification are from human society. Consequently, the system is unable to autonomously identify novel forms of privacy. Second, GAMA may be unable to fulfill the task successfully in instances where anonymization results in significant semantic loss. Future work will focus on two key areas. First, we will study how MAS autonomously learn laws and regulations, and establish new standards for privacy identification. Second, we will investigate multi-modal privacy-preserving MAS.

\begin{acks}
\end{acks}

\bibliographystyle{ACM-Reference-Format}
\bibliography{gama}

\appendix

\section{Knowledge Domain}
\label{app:knowledge domain}
From the perspective of academic disciplines, the classification of knowledge domains typically refers to a systematic and structured categorization based on established disciplinary frameworks within higher education systems, such as the CIP 2000 \footnote{CIP 2000: \href{https://nces.ed.gov/pubs2002/cip2000/}{https://nces.ed.gov/pubs2002/cip2000/}}. A clear hierarchical structure and well-defined disciplinary boundaries characterize this method of classification. It is widely applied in the design of higher education programs, research administration, talent development, and statistical analysis, serving as a foundational framework for organizing knowledge and constructing disciplinary systems. Therefore, we refer to the disciplinary classification in CIP2000 to categorize the task domains in our experiments as follows:
\begin{itemize}
    \item Entertainment
    \item Finance
    \item History
    \item Law
    \item Sociology
    \item Arts
    \item Literature
    \item Medicine
    \item Politics
    \item Sports
    \item Technology
    \item Business
    \item Communication
    \item Science
    \item Education
    \item Environment
    \item Agriculture
    \item Philosophy
    \item Transportation
    \item Psychology
\end{itemize}

\section{Llama3-8B Setting}
\label{app:llama}
Hyperparameters of Llama3-8B Model in the private space are as follows:
\begin{itemize}
    \item max\_new\_tokens: 256
    \item do\_sample:	True
    \item temperature: 0.6
    \item top\_p:	1.0
    \item return\_tensors: “pt”
    \item truncation:	True
    \item max\_length: 1024
\end{itemize}

\section{Metrics}
\label{app:metrics}
To ensure privacy preservation, we evaluate the performance of privacy identification using Precision, Recall, and F1-Score (F1). Additionally, we utilize BLEU \cite{papineni_bleu_2002} to assess the fluency of post-privacy-preserved questions and BERT-based Similarity (Similarity) \cite{tracz_bert-based_2020} to measure the semantic consistency between the original and post-privacy-preserved questions.
\begin{align}
    Precision &=\dfrac{TP}{(TP+FP)} \label{eq:precision}\\ 
    Recall &=\dfrac{TP}{(TP+FN)} \label{eq:recall}\\
    F1 &=\dfrac{2TP}{(2TP+FP+FN)} \label{eq:f1}\\
    Similarity&=\dfrac{BERT(x).BERT(y)}{\Vert BERT(x)\Vert \times \Vert BERT(y) \Vert}\\
    BLEU &=BP\ast\exp(\sum_{n=1}^{N} W_{n}\log(P_{n}))\label{eq:bleu}
\end{align}
where $TP$ represents True Positives, $TN$ represents True Negatives, $FP$ represents False Positives, $FN$ represents False Negatives, $BP$ denotes the penalty factor, and $N$ is the maximum token number of grams. $x$ is the prediction string and $y$ is the target string.In the original BLEU system, uniform weighting is adopted, $W_{n}=\frac{1}{N}$, and $P_n$ represents the precision of the $N$-gram.

Drawing on the approach of \cite{wang_unleashing_2024}, we adopt an automatic metric to identify factual errors and measure a model's capacity to integrate diverse domain knowledge. We conduct string matching with the veridical target answers for each question on the generated output. The generalized form is shown as Eq. (\ref{eq:score}).
\begin{equation}
    Score=\dfrac{A_{correct}}{N_q} \label{eq:score}
\end{equation}
Where $N_q$ is the number of questions, $A_{correct}$ is the number of correct answer mentions, Score is the metrics score for the tasks.

We evaluate the multi-agent system in action by analyzing the trajectories of the agents. For each trajectory $\mathcal{T}$, we prompt the language model (LM) to generate a final action, denoted as $a_\mathcal{T}$, to fulfill the corresponding user instruction. We then assess whether $a_\mathcal{T}$ leaks any information related to the data type defined in the seed $\mathcal{S}$.

To facilitate this evaluation, we extract a set of potentially sensitive information items from each trajectory, denoted as $\mathcal{I}(\mathcal{T}, \mathcal{S}) = {i_1, \dots, i_m}$, based on its seed $\mathcal{S}$. A few-shot LM-based classifier $f$ is constructed to determine whether each item $i_t$ can be inferred from the generated action $a_\mathcal{T}$. We consider $a_\mathcal{T}$ to leak private information if there exists any $t \in {1, \dots, m}$ such that $i_t$ is inferable from $a_\mathcal{T}$.

The leakage rate (LR) on the evaluation dataset $\mathcal{D}$ is defined as the proportion of actions $a_\mathcal{T}$ that leak at least one item in $\mathcal{I}(\mathcal{T}, \mathcal{S})$, as follow:

\begin{equation}
    LR=\frac{\sum_{(\mathcal{S,V,T})} { \{\cup f(i_t,a_{\mathcal{T}})|t=1,...,m\}}}{|\mathcal(D)|}
\end{equation}

\section{The details of datasets}
\label{app: dataset details}
\begin{itemize}
    \item \textbf{Trivia Creative Writing (TCW)} \cite{wang_unleashing_2024}.The task tests LLM's ability to retrieve and integrate diverse information from its internal knowledge. In this task, a model must craft a coherent story around a given topic while incorporating answers to N trivia questions. We evaluate the models with N set to 5 and 10, where a higher N requires more extensive domain knowledge. Our benchmark includes 100 instances for each N, totaling 1,000 trivia questions.
    \item \textbf{Logic Grid Puzzle (LGP)} \cite{wang_unleashing_2024}. The task is from the Bigbench dataset (Srivastava et al., 2023), which comprises 200 instances. Each instance describes a logic puzzle involving 2 to 5 houses, each occupied by a person with specific characteristics, such as playing the piano. The goal is to answer questions about house numbers based on given clues, requiring multi-step reasoning and the selection of relevant information. For evaluation, we measure the accuracy of the predicted house numbers by comparing them to the ground truth targets provided by the dataset.
    \item \textbf{Privacy Lens} \cite{shao_privacylens_2024}. It is a dataset released by Stanford University in 2023 to evaluate the behavior of large language models (LLMs) in privacy-related scenarios. It consists of approximately 600 manually crafted questions covering various types, such as direct inquiries, indirect reasoning, and social engineering-style prompts. The dataset is designed to assess whether a model refuses to answer, fabricates private information, or excessively discloses sensitive content. It evaluates model performance across multiple dimensions, including refusal behavior, hallucinated private information, and context sensitivity, aiming to measure the safety and alignment of LLMs when handling personally identifiable information. As a key benchmark for assessing privacy protection capabilities and the safety boundaries of models, Privacy Lens has been widely used to compare the privacy-related behaviors of models such as GPT-4, Claude, and PaLM.
\end{itemize}








\end{document}